\documentclass{ecai} 

\usepackage{latexsym}
\usepackage{amssymb}
\usepackage{amsmath}
\usepackage{amsthm}
\usepackage{booktabs}
\usepackage{enumitem}
\usepackage{graphicx}
\usepackage{color}

\usepackage{algorithm}
\usepackage{algorithmic}
\usepackage{subcaption}
\usepackage{multirow}
\usepackage{booktabs}
\usepackage{amssymb}

\newcommand{\BibTeX}{B\kern-.05em{\sc i\kern-.025em b}\kern-.08em\TeX}

\begin{document}


\begin{frontmatter}


\paperid{} 


\title{Boosting Adverse Weather Crowd Counting via Multi-queue Contrastive Learning}


\author[A]{\fnms{Tianhang}~\snm{Pan}
}
\author[A]{\fnms{Xiuyi}~\snm{Jia}
\thanks{Corresponding Author. }
}

\address[A]{School of Computer Science and Engineering, Nanjing University of Science and Technology}


\begin{abstract}
Currently, most crowd counting methods have outstanding performance under normal weather conditions.
However, our experimental validation reveals two key obstacles limiting the accuracy improvement of crowd counting models:
1) the domain gap bewteen the adverse weather and the normal weather images; 
2) the weather class imbalance in the training set. 
To address the problems, we propose a two-stage crowd counting method named Multi-queue Contrastive Learning (MQCL). 
Specifically, in the first stage, our target is to equip the backbone network with weather-awareness capabilities.
In this process, a contrastive learning method named multi-queue MoCo designed by us is employed to enable representation learning under weather class imbalance.
After the first stage is completed, the backbone model is ``mature'' enough to extract weather-related representations.
On this basis, we proceed to the second stage, in which we propose to refine the representations under the guidance of contrastive learning, enabling the conversion of the weather-aware representations to the normal weather domain.
Through such representation and conversion, the model achieves robust counting performance under both normal and adverse weather conditions.
Extensive experimental results show that, compared to the baseline, MQCL reduces the counting error under adverse weather conditions by 22\%, while introducing only about 13\% increase in computational burdon, which achieves state-of-the-art performance.
\end{abstract}

\end{frontmatter}


\section{Introduction}

\begin{figure}[tbp]
	\centering
    \includegraphics[width=240pt]{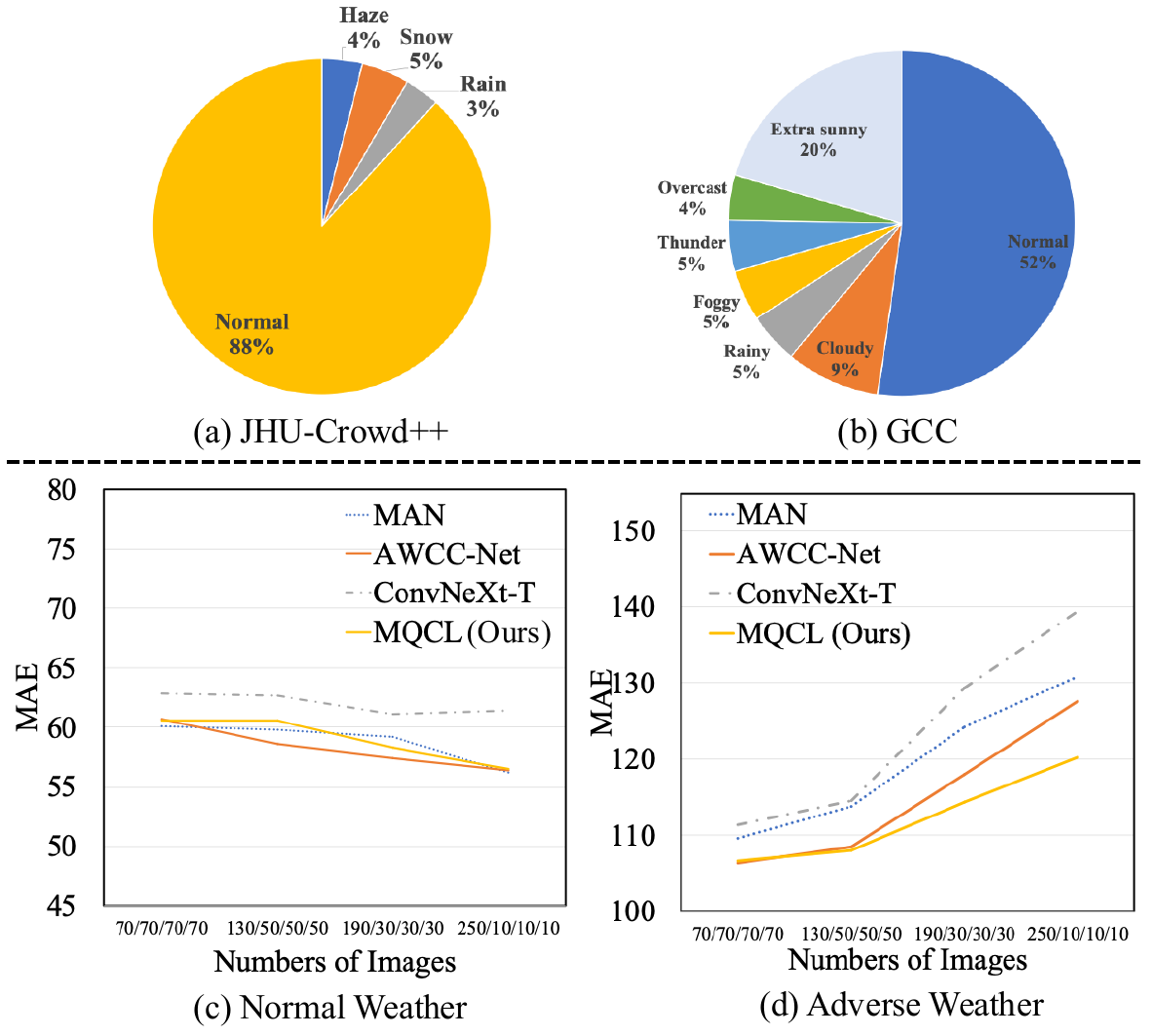}
	\caption[]{(a) and (b) shows the the weather condition distribution pie charts of the JHU-Crowd++ dataset and the GCC dataset, respectively. 
    (c) and (d) present the counting accuracy of different models on normal-weather and adverse-weather images respectively, under varying degrees of weather-class imbalance in the training data.
    The horizontal axis represents the number of normal/hazy/rainy/snowy weather images in the training set.
    The vertical axis represents the Mean Absolute Error (MAE) of models on the JHU-Crowd++ test set.
    All training data are sampled from the JHU-Crowd++ training set.
    }
    \vspace{15pt}
	\label{fig:figure1}
\end{figure}

Crowd counting has attracted much attention due to its wide range of applications such as public safety, video surveillance, and traffic control. 
Currently, most of the crowd counting methods~\cite{zhang2016single, li2018csrnet, lin2022boosting} are able to estimate the number of crowds well on the images recorded under normal weather conditions.
However, adverse weather such as haze, rain and snow can alter the texture and edge features of objects and even affect the clothing of people. 
The domain gap frequently results in counting difficulties.
The limited robustness to adverse weather fundamentally bottlenecks the advancement of crowd counting methods.
%
What's worse, labeled adverse weather crowd images are exceedingly rare.
For example, as shown in the pie charts in Figure~\ref{fig:figure1} (a), in the JHU-Crowd++ dataset~\cite{sindagi2020jhu}, the number of images under rain conditions accounts for only 3\% of the total dataset.
Such imbalance can severely damage model performance.
As shown in Figure~\ref{fig:figure1} (c) and (d), when training data imbalance increases, existing crowd counting models, such as MAN~\cite{lin2022boosting}, exhibit only negligible performance improvement on normal-weather images but severe degradation on adverse-weather images.

To mitigate the negative influence of adverse weather, an intuitive remedy is to pre-process the images using image restoration modules before counting.
%
Unfortunately, even when restoration modules can mitigate the obscuration caused by adverse weather, the restored images still exhibit significant domain differences from normal weather images.
For example, the restoration module cannot eliminate the occlusion caused by objects such as raincoats or umbrellas.
Moreover, the additionally introduced classification and enhancement modules significantly increase the computational burden. 
Training a separate model for each weather category is also not feasible because the limited samples of adverse weather conditions are insufficient to support the training. 
AWCC-Net~\cite{huang2023counting} is currently the only work that dives into how to improve model performance under diverse adverse weather conditions.
It introduces a transformer-based approach that addresses the issue of generating complementary information from image-specific degradation without the need for customized image enhancement modules.
However, its primary limitation is the disregard for performance degradation under weather-class imbalance.
As shown in Figure~\ref{fig:figure1} (d), it achieves significant performance improvement in adverse weather conditions, but still suffers severe degradation when class imbalance intensifies.
Moreover, the Transformer-based design poses a heavy computational burden to the model.

%
We aim to design a method that satisfies three objectives: (i) achieve high accuracy in adverse weather conditions even when training data is class-imbalanced; (ii) maintain high accuracy in normal weather conditions; (iii) be light-weight. 
To achieve above objectives, we approach the problem of crowd counting, which includes both normal and adverse weather conditions, as an imbalanced multi-domain learning task. 
The key prerequisite for a model to tackle multi-domain learning is to have perception capabilities across different domains, meaning that the extracted information from each domain should possess discriminative characteristics. 
This aligns with the objective of contrastive learning~\cite{chen2020simple, he2020momentum, oord2018representation}. 
In this paper, we propose a two-stage method called \textbf{M}ulti-\textbf{q}ueue \textbf{C}ontrastive \textbf{L}earning (\textbf{MQCL}). 
This approach enables the backbone model to directly extract weather-aware representations, which are further refined by a refiner module.
In the first stage, we use contrastive learning to distinguish the characteristics of different weather.
However, since the class imbalance mentioned above, normal weather samples dominate the key vector distribution compared to adverse weather samples during training, resulting in poor performance of representations and difficulty in refining and counting.
To tackle such problem, we design a simple yet effective contrastive learning method called multi-queue MoCo, which replaces the standard single queue in MoCo~\cite{he2020momentum} with multiple queues, providing class-balanced key vectors.
In the second stage, all normal weather samples in the queues are treated as positive samples of contrastive learning. 
Thus the refiner can be guided to convert the representations of adverse weather images to the domain of normal images. 
The following example can more concretely explain how MQCL works: 
Owing to the training during the WRL stage, the backbone model can encode humans with clear textures under normal weather as representation $A$, and humans wearing raincoats in rainy weather as representation $B$.
Furthermore, due to the training in the CRR stage, the refiner learns that within the context of crowd counting, representation $B$ is essentially equivalent to representation $A$, as both denote a single individual. 
It thus transforms representation $B$ into $A$.
Therefore, even under rainy conditions, the crowd counting head can still provide accurate count predictions.
As a plug-and-play method, MQCL perfectly satisfies the three aforementioned objectives: achieving high accuracy in both normal and adverse weather conditions, while demonstrating broad adaptability to diverse backbone architectures and high computational efficiency as only a light-weight refiner is introduced. 
As shown in Figure~\ref{fig:figure1} (c) and (d), MQCL achieves strong performance not only under balanced data conditions, but also demonstrates significantly greater robustness than other methods when faced with increased data imbalance.
The main contributions of our paper are concluded as follows.
\begin{itemize}
    \item To boost the robustness of the model under adverse weather conditions while maintaining normal-weather performance, we propose a light-weight two-stage method called MQCL, achieving significant improvement compared to the baseline. 
    \item To tackle the problem of class imbalance in contrastive learning, we propose a new method called multi-queue MoCo, achieving better performance than vanilla single-queue MoCo.
    \item To realize the conversion of representations from the adverse weather domain to the normal weather domain, we propose a refining method. Contrastive learning is utilized to guide the refinement process, enabling the decoder to focus on a single domain.  
    \item Extensive experimental results show that our method achieves competitive results. 
\end{itemize}

\section{Related Work}
\subsection{Crowd Counting under Normal Conditions}
Up to now, most single image crowd counting methods can be divided into two categories: density-map-based and localization-based crowd counting. 
Density-based methods aim to generate a density map, the pixel values of which represent the crowd density at that location.
The sum of all pixel values in the density map is the estimated total number of the image. 
MCNN~\cite{zhang2016single} is a pioneer in employing such a method. Benefiting from the multi-column design, MCNN can handle input images of arbitrary size or resolution.
CSRNet~\cite{li2018csrnet} employs dilated CNN for the back end to deliver larger reception fields and to avoid pooling operations. 
More recently, Lin et al. proposed a multifaceted attention network~\cite{lin2022boosting} to improve transformer models in local spatial relation encoding. 
Du et al. \cite{du2023redesigning} redesigns the multi-scale neural network by introducing a hierarchical mixture of density experts.
On the other hand, the target of localization-based methods is to predict head positions. 
Song et al.~\cite{song2021rethinking} propose a model called P2PNet to directly predict a set of point proposals to represent heads in an image, being consistent with the human annotation results.
They also proposed a new metric called density Normalized Average Precision (nAP) to provide more comprehensive and more precise performance evaluation. 
CLTR~\cite{liang2022end} is a concurrent work with P2PNet, introducing the Transformer architecture to the field of crowd localization and proposing a KMO-based Hungarian matcher to reduce the ambiguous points and generate more reasonable matching results. 
APGCC~\cite{chen2024improving} provides clear and effective guidance for proposal selection and optimization, addressing the
core issue of matching uncertainty in point-based crowd counting.

\subsection{Crowd Counting under Adverse Conditions}
Existing deep-learning-based methods have achieved unprecedented success with crowd counting, but their performance degraded severely under adverse conditions due to the disturbance to the brightness and gradient consistency. 
%
%
Additional class conditioning blocks are utilized by~\cite{sindagi2020jhu} to augment the backbone module, which is trained via cross-entropy error using labels available in the dataset. 
%
%
AWCC~\cite{huang2023counting} enables the model to extract weather information according to the degradation via learning adaptive query vectors, but the weight of the model is significantly increased due to the introduction of a Transformer-based module. 
Kong et al.~\cite{kong2023direction} proposes a single-stage hazy-weather crowd counting method based on direction-aware attention aggregation. 
RestoCrowd~\cite{kong2025restocrowd} is also a method focusing on haze-weather crowd counting, which leverages the image restoration coadjutant learning to achieve accurate counting without weather-label. 
However, these methods exhibit significant limitations, such as the inability to handle multiple types of unknown weather conditions or the neglect of class imbalance in weather categories within the dataset.
In this paper, MQCL not only be able to handle various types of unknown weather conditions, but also explicitly addresses the issue of weather category imbalance as one of its core design considerations.

\subsection{Contrastive Learning}
\label{sec:related_contrastive_learning}
Contrastive learning~\cite{oord2018representation, chen2020simple} has attracted much attention due to its success in unsupervised representation learning. 
The target of it is to maximize the similarity of the representations between positive pairs while minimizing that of negative pairs.
He et al.~\cite{he2020momentum} builds a dynamic dictionary with a queue and a moving-averaged encoder to enable large-scale contrastive learning with dramatically low demand for memory. 
\cite{khosla2020supervised} investigates the contrastive loss and adapted contrastive learning to the field of supervised learning. 
Despite the significant success achieved by contrastive learning, Assran et al.~\cite{assran2022hidden} revealed that in the formulation of contrastive methods is an overlooked prior to learn features that enable uniform clustering of the data.
This prior can hamper performance when pretraining on class-imbalanced data. 
To solve the class imbalance problem, Kang et al. proposed k-positive contrastive loss (KCL)~\cite{kang2020exploring}, which draws k instances from the same class to form the positive sample set. 
Balanced contrastive learning~\cite{zhu2022balanced} tries to mitigate the unbalanced problem from the aspects of re-weighting the gradient and ensuring batch data balance.
Hou et al.~\cite{hou2023subclass} clusters each head class into multiple subclasses of similar sizes as the tail classes and enforce representations to capture the two-layer class hierarchy between the original classes and their subclasses.

\section{Proposed Method}
In this work, we aim to improve the robustness of the model under multiple adverse weather conditions and maintain good performance under normal weather. 
Each sample in the training set consists of three components: the input image, the ground truth points of the human heads and a class label representing the class of the weather. 
Noting that the weather label is not available to the model in the inferencing phase, which requires the model to be able to deal with unknown corruptions. 
Images under adverse weather represent only a small part of the dataset. 
Thus, we formulate our problem as an imbalanced multi-domain learning problem. 

\subsection{Framework Overview}
\begin{figure}[tbp]
	\centering
	\includegraphics[width=250pt]{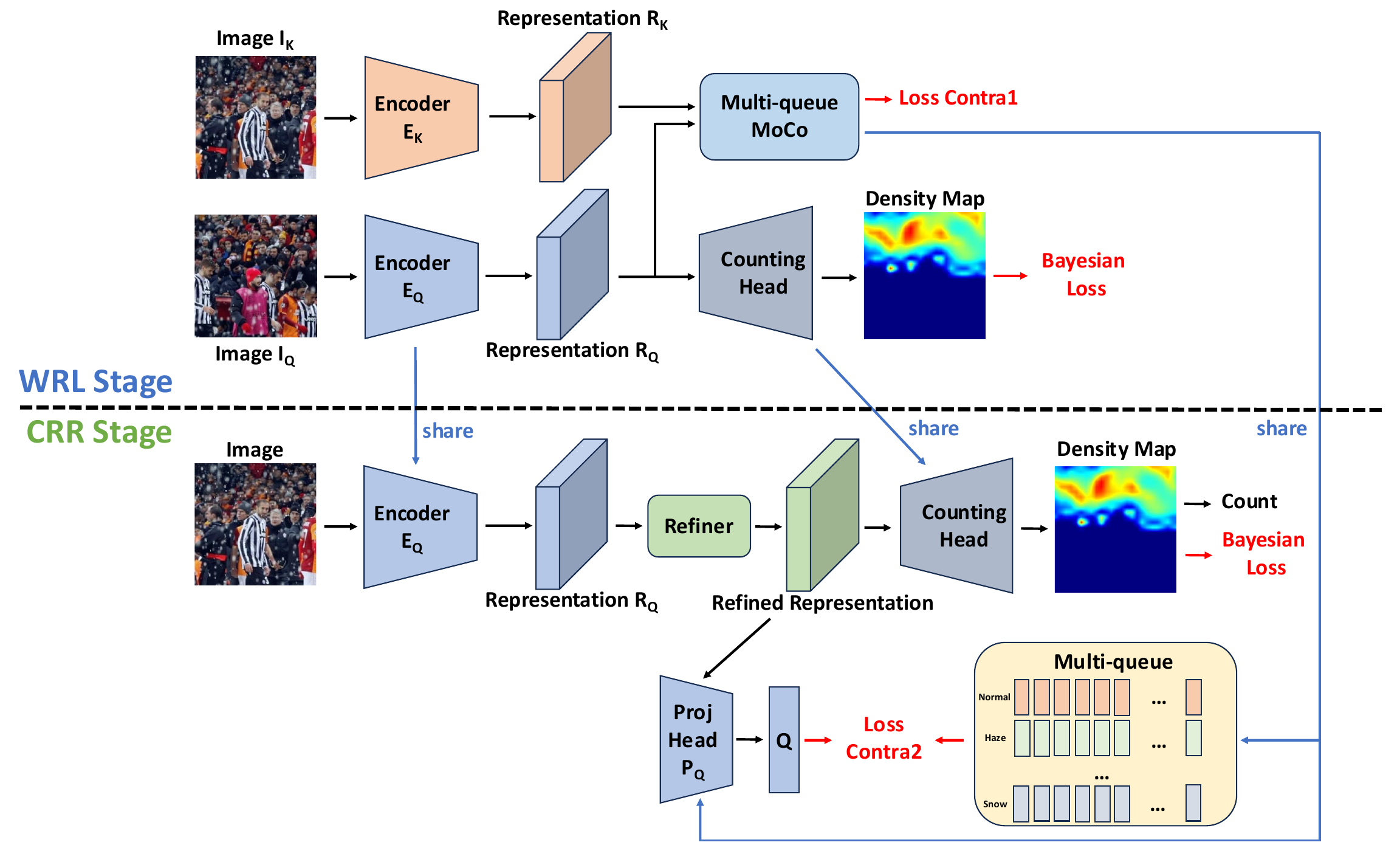}
	\caption{The architecture of MQCL.  
	The target of the WRL stage is to learn weather-aware representations via unsupervised contrastive learning.
    After the WRL stage, a refiner is trained in the CRR stage to pull the adverse weather representations towards normal weather domain.
	}
    \vspace{15pt}
	\label{fig:architecture}
\end{figure}

The architecture of our method is illustrated in Figure~\ref{fig:architecture}. 
As discussed earlier, we aim to directly enable the crowd counting backbone model to learn weather-aware feature representations and then refine it with a light-weight refiner. 
%
%
Since the refiner is premised on stable and consistent representation while it keeps evolving and is not stable during the representation learning stage, we divide the training into two stages to separate these two targets, namely Weather-aware Representation Learning (\textbf{WRL}) stage and Contrastive-learning-guided Representation Refining (\textbf{CRR}) stage, respectively. 
After training is completed, the model of the CRR stage is used for inference.
In the WRL stage, contrastive learning is utilized to enable the encoder to learn weather-aware representations.
The contrastive learning strategy of the WRL stage is based on our proposed multi-queue MoCo, which can tackle the problem of weather class imbalance. 
The weights of the encoder, the decoder and the queues obtained during the WRL stage will be retained for the CRR stage. 
In the CRR stage, the encoder is freezed and the main task of this stage is to train a refiner.  
By treating the normal weather samples as positive examples of contrastive learning, the refiner is guided to pull the adverse weather representations towards normal weather domain.
Finally, high-quality density map can be generated to realize precise counting. 
%
%
In terms of the counting head's structure, we remain consistent with MFA~\cite{ling2023motional}, which primarily consists of two transposed convolutional layers.

\subsection{Weather-aware Representation Learning}
The model architecture of the WRL stage is illustrated in the upper half of Figure~\ref{fig:architecture}. 
%
%
The contrastive learning method is utilized to train the encoders and endow it with the capability to extract weather-aware representations, i.e., images with similar weather conditions correspond to similar representations, whereas those with dissimilar conditions correspond to more distant representations. 
Our training structure is similar to that of MoCo~\cite{he2020momentum}, which consists of an encoder $\rm E_Q$, a momentum-updated encoder $\rm E_K$ and a crowd countign head. 
The encoders $\rm E_Q$ and $\rm E_K$ extracts representations $\rm R_Q$ and $\rm R_K$, respectively, from different augmentations $\rm I_Q$ and $\rm I_K$ of an image. 
The representations $\rm R_Q$ and $\rm R_K$ will be further processed by our proposed multi-queue MoCo, which tackles the problem of weather class imbalance and grant the representations with weather-aware nature, to obtain the contrastive loss.
The details of multi-queue MoCo will be elaborated in Section~\ref{sec:mq_moco}.
To ensure that the representation simultaneously contains crowd information, the target of the decoder is set as generating a density map under the supervision of the Bayesian loss \cite{ma2019bayesian} according to the representation $\rm R_Q$. 
The total loss of the WRL stage is: 
\begin{equation}
	\mathcal{L}_{wrl} = \mathcal{L}_{contra1} + \lambda_1 \mathcal{L}_{bayesian}, 
\end{equation}
where $\mathcal{L}_{contra1}$ is the contrastive loss based on the multi-queue MoCo and $\mathcal{L}_{bayesian}$ is the Bayesian loss. 

\subsection{Multi-queue MoCo}
\label{sec:mq_moco}
Theoretical reasoning and experimental evidence in~\cite{assran2022hidden} suggest that contrastive learning has an overlooked prior-to-learn feature that enables uniform clustering of the data and it can hamper performance when training on class-imbalanced data. 
In the vanilla contrastive learning strategy, positive and negative samples are entirely obtained through random sampling. 
This strategy can work perfectly under class-balanced conditions but may struggle if the data is imbalanced due to the mismatch between the actual distribution and the model's prior. 

\begin{figure}[tbp]
	\centering
	\includegraphics[width=230pt]{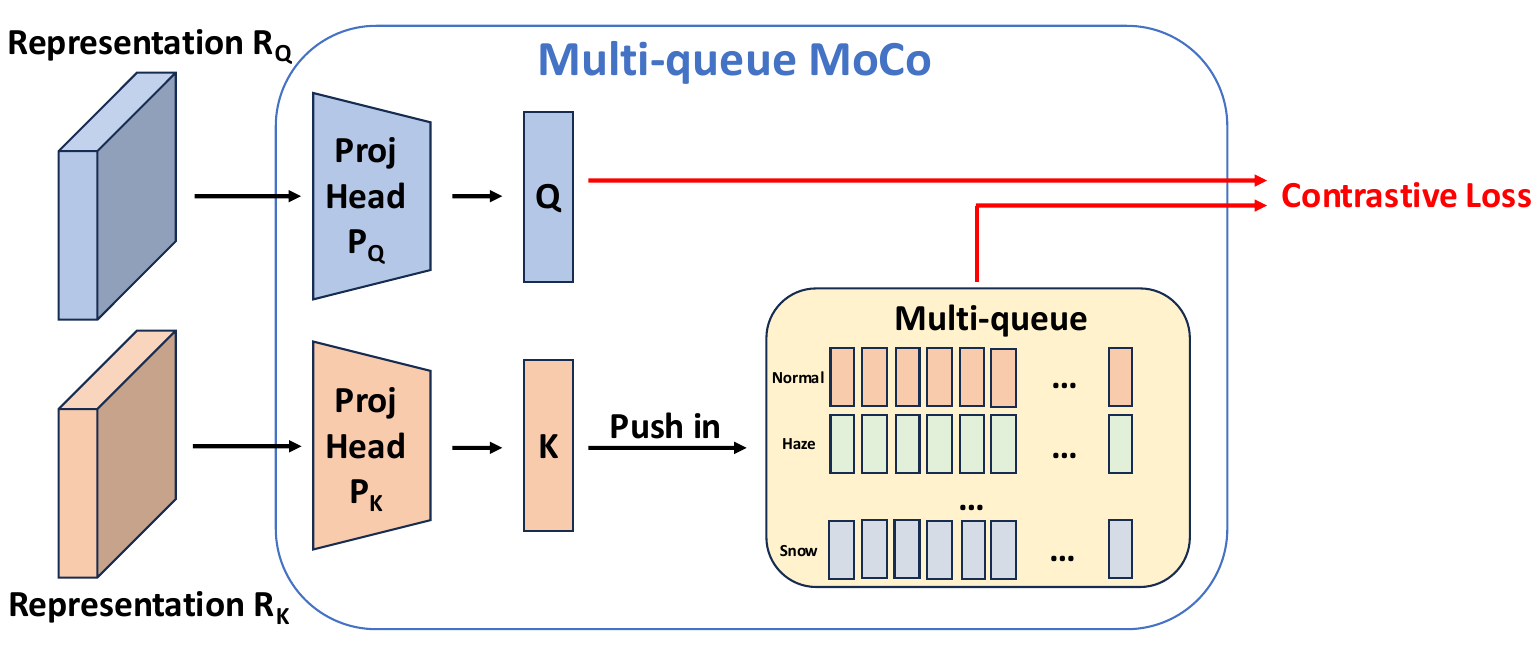}
	\caption{The architecture of multi-queue MoCo. The projection heads project the representations to 1-D vectors. In the multiple queues, each sub queue is of equal length and corresponds to one weather class.}
    \vspace{15pt}
	\label{fig:multi_queue_moco}
\end{figure}

To tackle the problem, we propose multi-queue MoCo, the architecture of which is illustrated in Figure~\ref{fig:multi_queue_moco}. 
Similar to most of the contrastive methods, each image undergoes data augmentation multiple times, and the representations originating from the same image as the anchor are treated as positive samples while those from different images are treated as negative samples. 
This strategy may lead to a situation where samples with same weather conditions from different images are treated as negatives.
This seems to be inconsistent with our motivation.
But fortunately, research conducted by~\cite{wang2021understanding} indicates that contrastive learning has the tolerance to semantically similar negative samples.
In light of this, we conducted extensive experiments, which are presented in Section~\ref{sec:key_issues}, and the results indicate that such a strategy outperforms the positive/negative partitioning strategy based on weather labels. 
Both representations $\rm R_Q$ and $\rm R_K$ can be regarded as tensors of $\mathbb{R}^{H \times W \times C_1}$, where $H$, $W$ and $C_1$ are the height, the width and the number of channels of the representation, respectively. 
To avoid information loss introduced by the contrastive loss and reduce computational complexity, nonlinear projection heads~\cite{chen2020simple} are introduced after encoders to project the representations to 1-D vectors. 
The projection head first pools the representations to vectors of $\mathbb{R}^{C_1}$ and then project them to vectors Q or K of $\mathbb{R}^{C_2}$ by introducing a multi-layer perceptron. $C_2$ is the dimension of the vectors. 
We refer to the projection heads following encoder $\rm E_Q$ and $\rm E_K$ as projection head $\rm P_Q$ and $\rm P_K$, respectively. 
They have the same structure but do not share parameters with each other.
In contrast to MoCo, to achieve a uniform distribution of classes within vectors in the memory, we improve the original single queue to a multi-queue structure. 
The number of sub queues in the multiple queues is equal to the number of classes, with each sub queue having an equal length and exclusively storing vectors Q that match its corresponding class.
The multi-queue structure can be considered as a tensor of $\mathbb{R}^{B \times L \times C_2}$, where $B$ is the number of classes and $L$ is the length of each sub queue. 
Immediately when the computation of vector K is completed, it will be pushed into the corresponding sub queue according to its weather class label.
With this design, the number of samples of each weather class in the memory becomes equal, aligning perfectly with the uniform prior of the contrastive loss.
Moreover, due to the limited number of images from adverse weather in the dataset, multiple samples from the same image may coexist within a sub queue. 
To avoid treating the above samples negative to each other, we propose to assign a unique index value to each image in the dataset and treat the samples corresponding to the same index as positive. 
The loss function of our multi-queue MoCo can be calculated as follows:
\begin{equation}
	\mathcal{L}_{contra1} = \sum_{i \in I} \frac{-1}{|P(i)|} \sum_{p \in P(i)} \log \frac{\exp({\rm Q_i} \cdot {\rm K_p} / \tau)}{\sum_{a \in \mathbf{A}} \exp({\rm Q_i} \cdot {\rm K_a} / \tau)}, 
\end{equation}
where $I$ is the batch size, $\rm Q_i$ is the anchor vector, $P(i)$ is the set of indices of the vectors K originating from the same image with $\rm Q_i$, $\mathbf{A}$ is the set of the indices of all of the vectors in the multiple queues and $\tau$ is the temperature. 

\subsection{Contrastive-learning-guided Representation Refining}
After the WRL stage, we can assume that the encoder is ``mature'' enough to effectively extract the weather and crowd information from images. 
The task of the CRR stage is to train a refiner which can convert the weather-aware representations to the normal weather domain, generating representations of the same size of $\mathbb{R}^{H \times W \times C_1}$ and enabling the decoder to focus on a single domain. 
The multiple queues obtained during the WRL stage have stored a sufficient number of vectors from various weather conditions.
In this paper, we propose to preserve and freeze these queues and employ contrastive learning to guide the refiner to convert the representation. 
In this stage, all the vectors in the normal-weather sub queue are treated as positive samples and those in other sub queues are treated as negatives. 
Since there is no longer a need to generate vectors K, the encoder $\rm E_K$ and projection head $\rm P_K$ are discarded. 
In order to maintain stable representations, the parameters of encoder $\rm E_Q$ are fixed at this stage. 
The projection head $\rm P_Q$ is preserved, fixed, and moved behind the refiner. 
We denote the vector projected from the refined representation as $\rm Q^{\prime}$.
The model structure of the CRR stage is shown in the lower half of Figure~\ref{fig:architecture}. 
The loss function of contrastive learning in this stage is calculated as follows:
\begin{equation}
	\mathcal{L}_{contra2} = \sum_{i \in I} \frac{-1}{|\mathbf{N}|} \sum_{p \in \mathbf{N}} \log \frac{\exp(\rm Q^{\prime}_i \cdot {\rm K_p} / \tau)}{\sum_{a \in \mathbf{A}} \exp({\rm Q^{\prime}_i} \cdot {\rm K_a} / \tau)}, 
\end{equation}
where $\rm N$ is the set of the indices of the vectors K in the normal-weather sub queue. 
The main difference between the $\mathcal{L}_{contra1}$ and $\mathcal{L}_{contra2}$ is that $\mathcal{L}_{contra1}$ treats samples originate from the same image as positive examples, while $\mathcal{L}_{contra2}$ treats all the normal weather samples in the retained multiple queues as positive examples. 
Similar to the WRL stage, the CRR stage continues to utilize the Bayesian loss to supervise the density map.
The overall loss function $\mathcal{L}_{CRR}$ for this stage is calculated as follows:
\begin{equation}
	\mathcal{L}_{CRR} = \mathcal{L}_{contra2} + \lambda_2 \mathcal{L}_{bayesian}, 
\end{equation}
where $\mathcal{L}_{contra2}$ is the contrastive loss and $\mathcal{L}_{bayesian}$ is the Bayesian loss.  

\section{Experiment and Discussion}
\subsection{Experiment Setup}
In MQCL, multiple possible architectural configurations can be used as Encoder $\rm E_Q$ and $\rm E_K$.
%
We employ four distinct models as encoders in our experiments: VGG-16~\cite{simonyan2014very}, CSRNet~\cite{li2018csrnet}, Swin-T~\cite{liu2021swin} and ConvNeXt-T~\cite{liu2022convnet}.
Except for the removal of the classification head, the architecture of encoders remain identical to the original implementation.
%
%
The structure of the crowd counting head is consistent with MFA~\cite{ling2023motional}, which is an upsampling module composed of two transposed convolutional layers.
Two-layer MLPs are employed in the projection heads, the output dimension of which is 2048 and 128. 
The refiner consists of three ConvNeXt blocks with input and output dimensions of 768, initialized with random parameters. 
As mentioned above, each image in a batch undergoes augmentation twice.
Specifically, we random crop the image with a size of 256 $\times$ 256, and horizontal flipping is applied for a probability of 50\%. 
In the multi-queue structure, the number of the sub queues is equal to the number of weather classes, and the length of each sub queue is set to 1024. 
The AdamW optimizer~\cite{loshchilov2017decoupled} is adopted both in the WRL and the CRR stage, the learning rate is scheduled by a cosine annealing strategy and the initial learning rate is $10^{-4}$. 
The weight decay is set to $10^{-3}$ and the batch size is $16$. 
$\lambda_1$ and $\lambda_2$ in the loss function are both set to 10 and the temperature $\tau$ is set to 0.05. 
Following the convention of existing works~\cite{li2018csrnet, lin2022boosting}, we adopt Mean Absolute Error (MAE) and Root Mean Squared Error (RMSE) as the metrics to evaluate the methods.

\begin{table}[tbp]
	\centering
	\tiny
    \caption{Quantitative results comparing with the state-of-the-art methods on the JHU-Crowd++ dataset. The best performance is shown in \textbf{bold} and the second best is shown in \underline{underlined}.}
    \vspace{5pt}
	\begin{tabular}[]{ccccccc}
		\toprule
		\multirow{2}{*}{Method} & \multicolumn{2}{c}{Normal} & \multicolumn{2}{c}{Adverse} & \multicolumn{2}{c}{Total} \\
		& MAE $\downarrow$ & RMSE $\downarrow$ & MAE $\downarrow$ & RMSE $\downarrow$ & MAE $\downarrow$ & RMSE $\downarrow$ \\
		\midrule
		SFCN~\cite{wang2019learning} (CVPR 19) & 71.4 & 225.3 & 122.8 & 606.3 & 77.5 & 297.6 \\
		BL~\cite{ma2019bayesian} (ICCV 19) & 66.2 & 200.6 & 140.1 & 675.7 & 75.0 & 299.9 \\
		LSCCNN~\cite{sam2020locate} (PAMI 20) & 103.8 & 399.2 & 178.0 & 744.3 & 112.7 & 454.4 \\
		CG-DRCN-V~\cite{sindagi2020jhu} (PAMI 20) & 74.7 & 253.4 & 138.6 & 654.0 & 82.3 & 328.0 \\
		CG-DRCN-R~\cite{sindagi2020jhu} (PAMI 20) & 64.4 & 205.9 & 	120.0 & 580.8 & 71.0 & 278.6 \\
		UOT~\cite{ma2021learning} (AAAI 21) & 53.1 & 148.2 & 114.9 & 610.7 & 60.5 & 252.7 \\
		GL~\cite{wan2021generalized} (CVPR 21) & 54.2 & 159.8 & 115.9 & 602.1 & 61.6 & 256.5 \\
		CLTR~\cite{liang2022end} (ECCV 22) & 52.7 & \underline{148.1} & 109.5 & 568.5 & 59.5 & 240.6 \\
		MAN~\cite{lin2022boosting} (CVPR 22) & \textbf{46.5} & \textbf{137.9} & 105.3 & 478.4 & 53.4 & 209.9 \\
		AWCC-Net~\cite{huang2023counting} (ICCV 23) & 47.6 & 153.9 & 87.3 & \textbf{430.1} & \underline{52.3} & \underline{207.2} \\
        LDFNet~\cite{chen2024learning} (TIP 24) & 52.8 & 156.5 & 112.4 & 530.1 & 59.9 & 234.8 \\
        APGCC~\cite{chen2024improving} (ECCV 24) & 48.4 & 169.3 & 108.2 & 478.4 & 55.5 & 229.3 \\
        Gramformer~\cite{lin2024gramformer} (AAAI 24) & \underline{46.7} & 170.2 & 101.2 & 471.4 & 53.1 & 228.1 \\
		\midrule
        VGG-16~\cite{simonyan2014very} & 68.2 & 221.4 & 115.3 & 632.0 & 73.8 & 301.4 \\
        MQCL + VGG-16 & 67.2 & 209.2 & 101.4 & 533.4 & 71.3 & 269.3 \\
        CSRNet~\cite{li2018csrnet} & 64.3 & 223.5 & 114.8 & 603.9 & 70.3 & 295.8 \\
        MQCL + CSRNet & 63.1 & 207.7 & 99.8 & 540.2 & 67.5 & 269.9 \\
        Swin-T~\cite{liu2021swin} & 57.8 & 171.8 & 103.5 & 471.7 & 63.3 & 229.2 \\
        MQCL + Swin-T & 50.6 & 163.8 & \underline{85.5} & 458.8 & 54.8 & 220.8 \\
		ConvNeXt-T~\cite{liu2022convnet} (CVPR 22) & 52.7 & 154.9 & 105.1 & 561.4 & 59.0 & 242.4 \\
        MultiModel + ConvNeXt-T & 52.9 & 157.0 & 108.2 & 542.2 & 59.5 & 238.3 \\
        Restore + ConvNeXt-T & 52.7 & 155.9 & 99.4 & 533.1 & 58.3 & 235.2 \\
		MQCL + ConvNeXt-T & 48.1 & 150.0 & \textbf{82.2} & \underline{433.5} & \textbf{52.1} & \textbf{205.5} \\
		\bottomrule
	\end{tabular}
	\label{tbl:jhu_performance}
\end{table}

\subsection{JHU-Crowd++ Dataset}
\begin{figure*}[tbp]
	\centering
	\includegraphics[width=500pt]{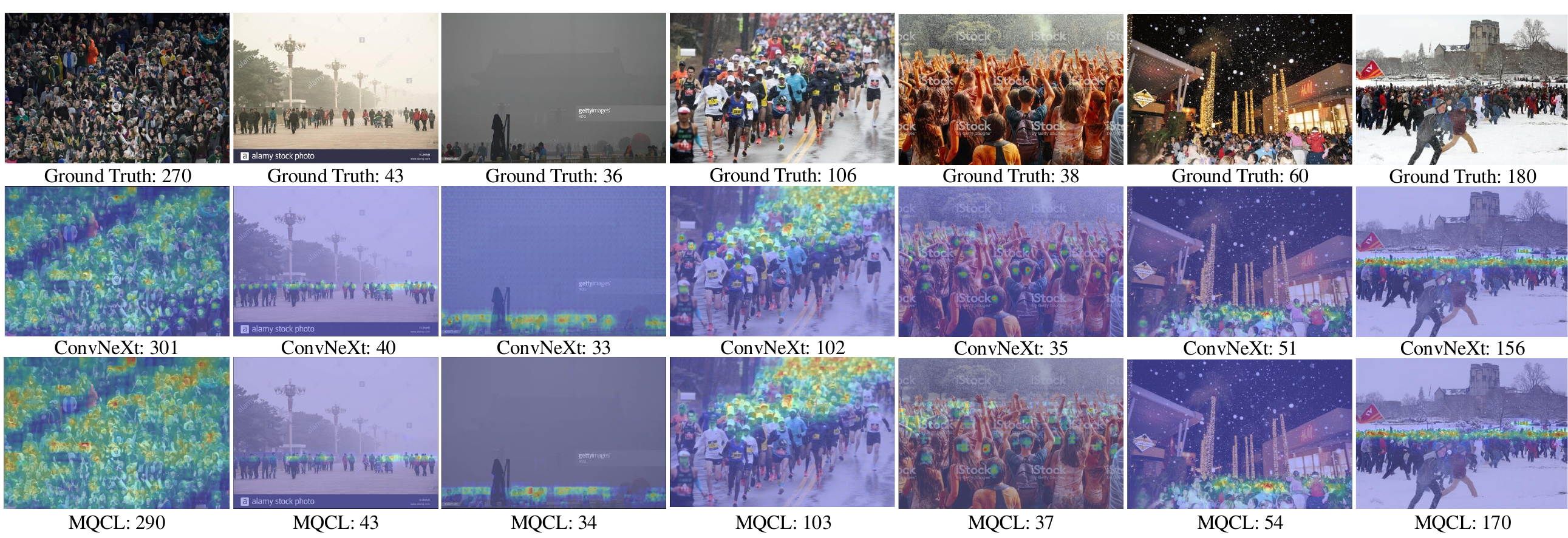}
    \centering
	\caption{Visualizations of our baseline ConvNeXt-T and our MQCL. The first row are the input images. The second and the third rows are the density maps predicted by ConvNeXt and our method, respectively. The images are sampled from the JHU dataset, including weather conditions of normal, haze, rain and snow. }
    \vspace{10pt}
	\label{fig:visualization}
\end{figure*}

There are 4372 images and 1.51 million labels contained in the JHU-Crowd++ dataset~\cite{sindagi2020jhu}. 
Out of these, 2272 images were used for training, 500 images for validation, and the remaining 1600 images for testing.
The advantage of JHU-Crowd++ is its inclusion of diverse scenes and environmental conditions, such as rain, snow and haze. 
It also provides weather condition labels for each image. 
Due to the rarity of adverse weather, the weather classes in JHU-Crowd++ are imbalanced.
As is shown in Figure~\ref{fig:figure1} (a), the number of images under rain, snow, and haze conditions accounts for only 3\%, 5\%, and 4\% of the total dataset, respectively.
Table~\ref{tbl:jhu_performance} shows the performance of MQCL and other state-of-the-art methods on the JHU-Crowd++ dataset.
VGG-16, Swin-T and ConvNeXt-T in the table stand for the baseline models, which have the same structure with their original image classification models, except for replacing the classification heads with crowd counting heads. 
Despite the challenges posed by the diverse scenes, complex and variable weather conditions and weather class imbalance in the JHU-Crowd++ dataset, MQCL achieves significant performance improvements regardless of the backbone used.
For example, MQCL achieves an improvement of 21.8\% in MAE and 22.8\% in RMSE under adverse weather conditions compared to the baseline ConvNeXt-T.
Moreover, MQCL can also improve the performance under normal conditions, decreasing MAE and RMSE by 8.7\% and 3.2\% compared to baseline ConvNeXt-T, respectively.
We believe that this can be attributed to the contrastive learning strategy employed in this paper, which considers augmentations from the same image as positive examples. 
This strategy not only assists the model in weather perception but also strengthens the model's ability to recognize different scenes. 
Sampling from the same image ensures that the positive examples not only share the same weather conditions but also possess similar scene characteristics.
On the other hand, owing to the superior performance in adverse weather conditions, MQCL surpasses state-of-the-art models such as MAN and AWCC-Net in overall test-set performance.
In addition, we conducted experimental verification of two simple methods mentioned in the Introduction, both based on the ConvNeXt-T baseline. 
The approach of training a separate model for each weather condition corresponds to ``MultiModel'' in the table. 
For this method, ResNet-50~\cite{he2016deep} is used as the weather classifier. 
It is evident that the limited number of adverse weather images in the training set is insufficient to support the training of high-performance models, leading to worse performance under adverse weather compared to the baseline. 
Moreover, occasional misclassifications by the weather classifier also negatively impact performance under normal weather conditions.
The approach of performing image restoration before counting corresponds to ``Restore'' in the table, where we adopt the Unified model~\cite{chen2022learning} for restoration. 
Under this method, the performance under adverse weather conditions shows some improvement, but the gains are significantly smaller compared to those achieved by MQCL.
The visualizations of ConvNeXt-T and our MQCL are compared in Figure~\ref{fig:visualization}.

\subsection{GCC Dataset}
GTA5 Crowd Counting (GCC) dataset~\cite{wang2019learning} is a fully synthetic large scale dataset, the scenarios of which is derived from the computer game Grand Theft Auto V (GTA5). 
The dataset was initially proposed for pretraining the model and boosting the performance on the real-world datasets by finetuning or adaptation. 
Here, we use it as a benchmark to validate the performance of crowd counting models under adverse weather conditions on large-scale data.
There are 15212 images and 7.63 million labels contained in the GCC dataset. 
Since each location in the dataset encompasses multiple cameras, \cite{wang2019learning} provides three strategies for splitting the training and the testing set: random, cross-camera and cross-location. 
Here we choose the cross-location strategy, which means 75\%/25\% locations are utilized for training/testing, since this scheme is most effective in validating the robustness of the model against unknown scenes. 
The weather conditions in the GCC dataset are diverse, comprising a total of seven types: normal, cloudy, rainy, foggy, thunder, overcast and extra sunny. 
They are evidently unevenly distributed either and the pie chart of the weather distribution is shown in Figure~\ref{fig:figure1} (b).
In our implementation, we randomly selected 10 locations out of the 75 in the training set as a validation set.

\begin{table}[tbp]
	\centering
	\tiny
    \caption{Quantitative results comparing with the state-of-the-art methods on the GCC dataset. The best performance is shown in \textbf{bold} and the second best is shown in \underline{underlined}. }
    \vspace{5pt}
	\begin{tabular}[]{ccccccc}
		\toprule
		\multirow{2}{*}{Method} & \multicolumn{2}{c}{Normal} & \multicolumn{2}{c}{Adverse} & \multicolumn{2}{c}{Total} \\
		& MAE $\downarrow$ & RMSE $\downarrow$ & MAE $\downarrow$ & RMSE $\downarrow$ & MAE $\downarrow$ & RMSE $\downarrow$ \\
		\midrule
		CSRNet~\cite{li2018csrnet} (CVPR 18) & 64.8 & 180.5 & 87.4 & \underline{220.8} & 75.5 & \underline{200.7} \\
		BL~\cite{ma2019bayesian} (ICCV 19) & 104.5 & 252.6 & 137.2 & 304.0 & 120.0 & 278.2 \\ 
		DM-Count~\cite{wang2020distribution} (NeurIPS 20) & 78.4 & 207.6 & 103.5 & 248.5 & 90.3 & 227.9 \\
		KDMG~\cite{wan2020kernel} (PAMI 20) & 94.6 & 225.0 & 124.0 & 280.5 & 108.6 & 252.9 \\
		MAN~\cite{lin2022boosting} (CVPR 22) & 70.1 & 207.8 & 94.5 & 246.4 & 81.7 & 227.0 \\
        LDFNet~\cite{chen2024learning} (TIP 24) & \underline{60.1} & 182.8 & \underline{80.1} & 243.7 & \underline{69.6} & 213.9 \\
        APGCC~\cite{chen2024improving} (ECCV 24) & 63.2 & \underline{175.9} & 87.2 & 258.5 & 74.6 & 219.1 \\
		\midrule
		ConvNeXt-T~\cite{liu2022convnet} (CVPR 22) & 65.3 & 184.5 & 85.5 & 221.7 & 74.9 & 203.0 \\
		MQCL + ConvNeXt-T & \textbf{59.7} & \textbf{175.5} & \textbf{79.7} & \textbf{210.5} & \textbf{69.2} & \textbf{192.9} \\
		\bottomrule
	\end{tabular}
	\label{tbl:gcc_performance}
\end{table}

We conducted quantitative experiments on the GCC dataset using MQCL with a ConvNeXt-T backbone.
Note that fewer comparison methods is included on this dataset because all performance results are obtained from our own training, as shown in Table~\ref{tbl:gcc_performance}.
Our MQCL achieves optimal performance under both normal and adverse weather conditions on all the metrics. 
Compared to our baseline ConvNeXt-T, MQCL decreases the MAE/RMSE under normal and adverse weather conditions by 8.6\%/4.9\% and 6.8\%/5.1\%, respectively. 
This indicates that our MQCL can also outperforms other methods when trained on relatively large-scale datasets and more importantly, our multi-queue MoCo strategy performs well even in the case of higher number of sub queues (7 for example).

\subsection{Key Issues and Discussions}
\label{sec:key_issues}

\begin{table}[tbp]
	\centering
	\tiny
    \caption{Performance comparison of the two optional strategies on the JHU-Crowd++ dataset. Strategy 1 treats samples with the same weather label as positive examples and strategy 2 treats samples originating from the same image as positive examples.}
    \vspace{5pt}
	\begin{tabular}[]{ccccccc}
		\toprule
		\multirow{2}{*}{Method} & \multicolumn{2}{c}{Normal} & \multicolumn{2}{c}{Adverse} & \multicolumn{2}{c}{Total} \\
		& MAE $\downarrow$ & RMSE $\downarrow$ & MAE $\downarrow$ & RMSE $\downarrow$ & MAE $\downarrow$ & RMSE $\downarrow$ \\
		\midrule
		Strategy 1 & 51.4 & 153.0 & 89.7 & 470.4 & 56.0 & 216.9 \\
		Strategy 2 (Ours) & \textbf{48.1} & \textbf{150.0} & \textbf{82.2} & \textbf{433.5} & \textbf{52.1} & \textbf{205.5}  \\
		\bottomrule
	\end{tabular}
    \vspace{10pt}
	\label{tbl:strategy_comparison}
\end{table}

\textbf{Strategy of positive/negative selection. }
As mentioned above, there are two optional strategies in the WRL stage: 
1) treating samples with the same weather label as positive examples;
%
2) treating samples originating from the same image as positive examples;
%
We conduct experiments on the JHU-Crowd++ dataset to compare these two strategies. 
The model performance under these two strategies is shown in Table~\ref{tbl:strategy_comparison}. 
Strategy 2 outperforms Strategy 1 in both normal and adverse conditions.
We attribute this to the following reasons: 
1) even if two samples share the same weather label, their weather conditions may still vary significantly.
The practice of minimizing all the representations with the same label is not in line with the target of us; 
2) samples from the same image not only share the same weather conditions but also the same scene.
Strategy 2 has the potential to enhance the model's scene recognition capabilities.

\begin{figure*}[tbp]
	\centering
    \subcaptionbox{\label{1}}{
        \includegraphics[height=150pt]{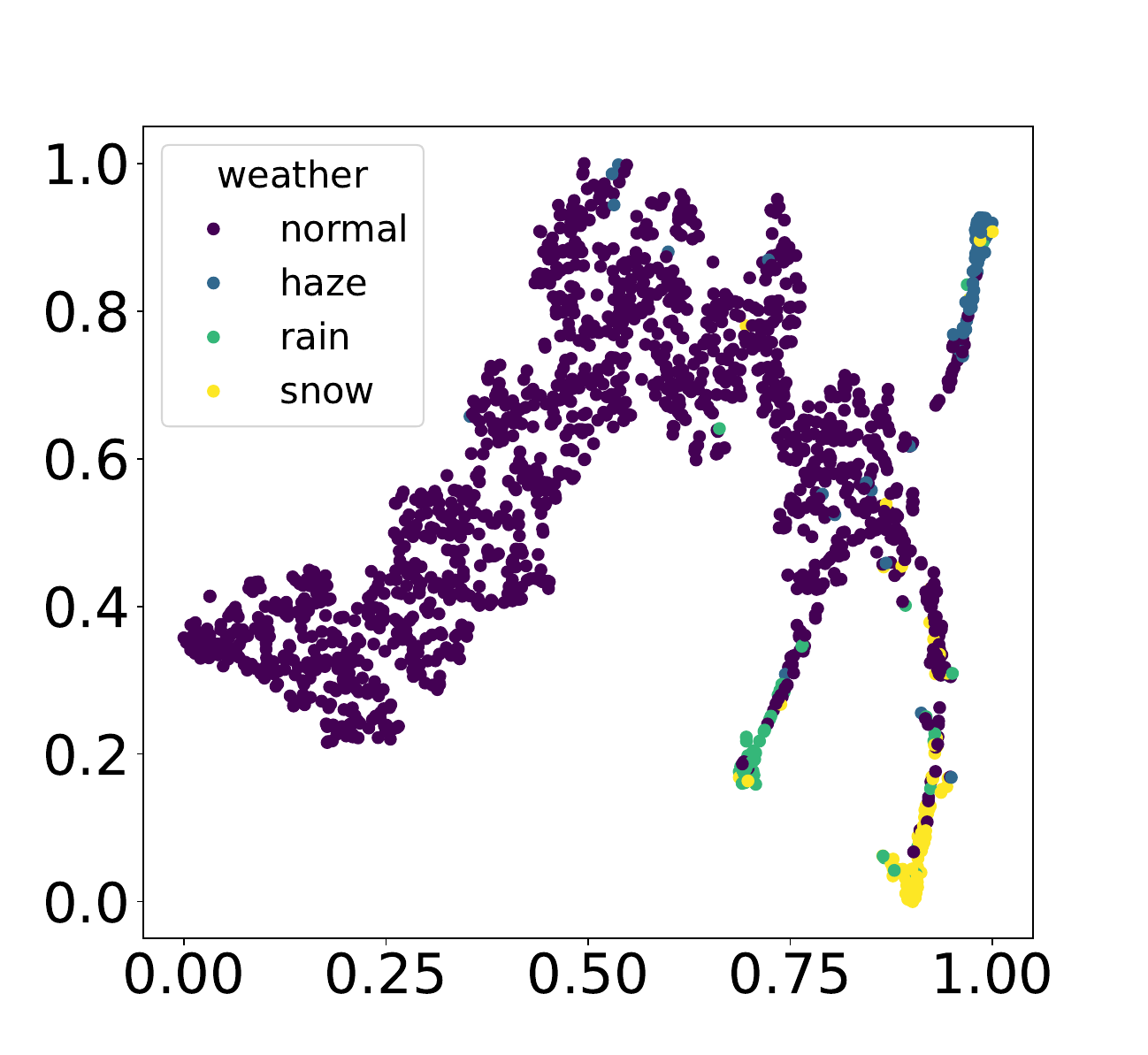}
    }
    \subcaptionbox{\label{2}}{
        \includegraphics[height=150pt]{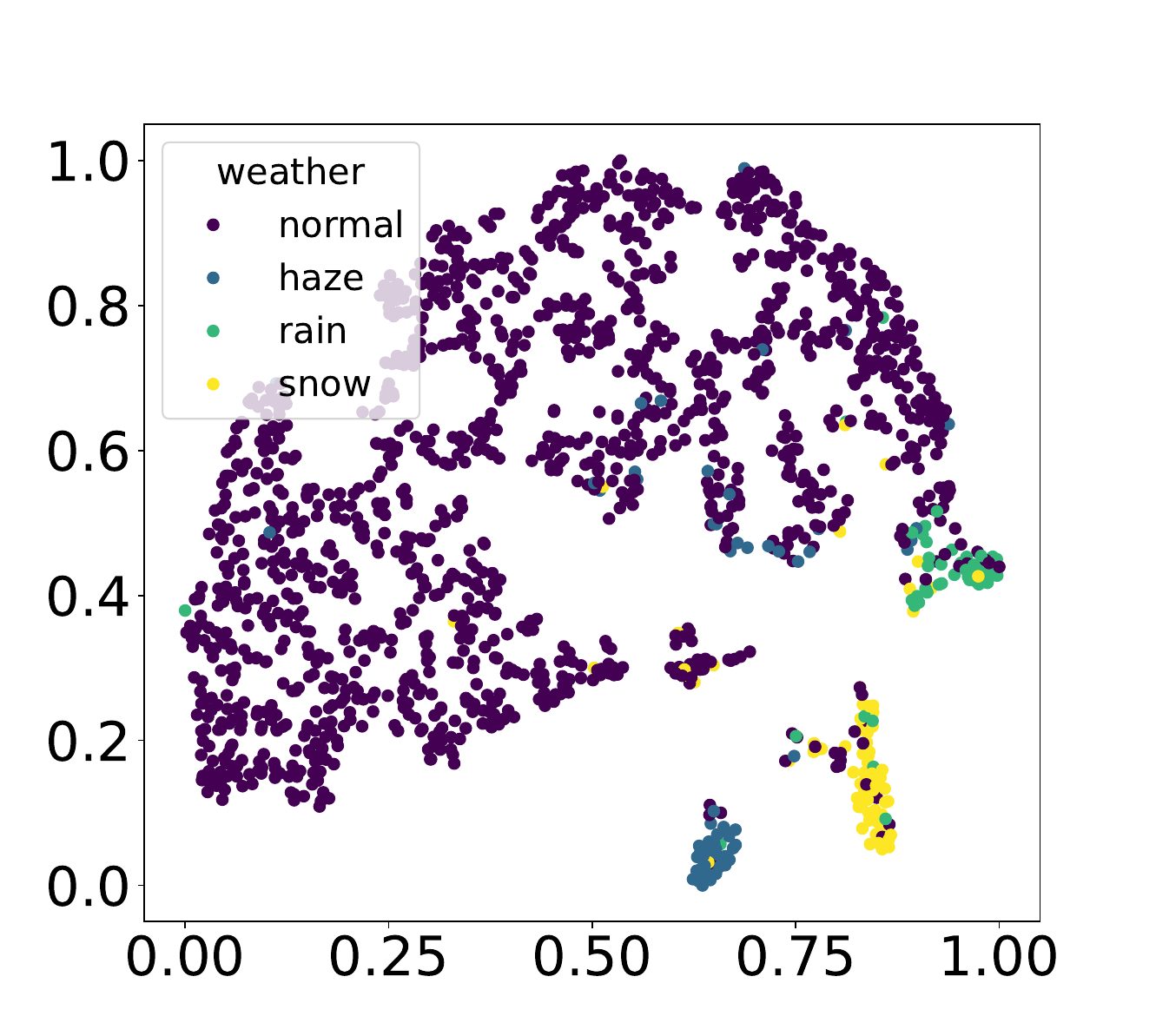}
    }
    \subcaptionbox{\label{3}}{
        \includegraphics[height=150pt]{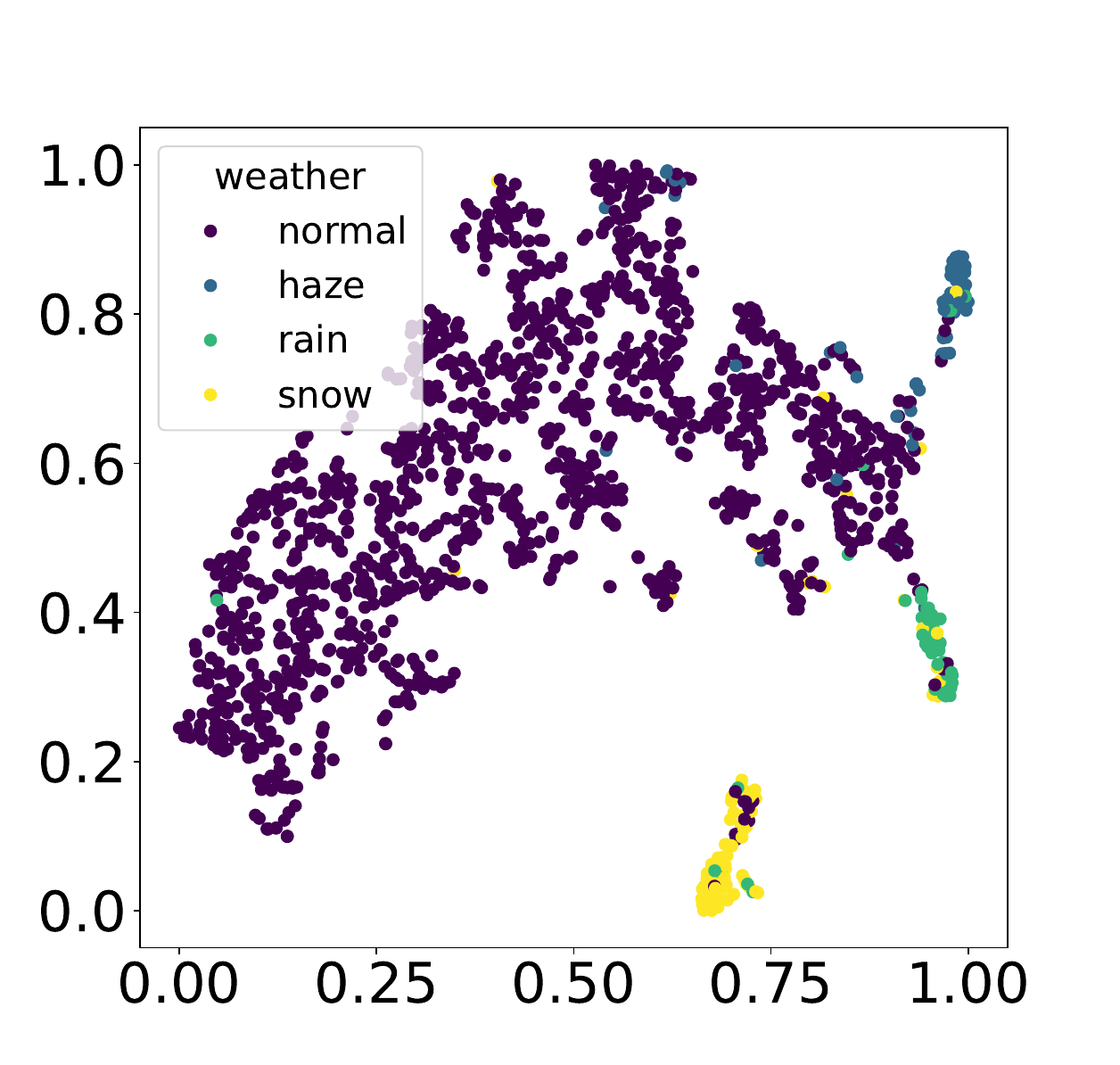}
    }
    \vspace{10pt}
	\caption[]{The t-SNE~\cite{van2008visualizing} visualization of the vectors Q after the WRL stage on the JHU-Crowd++ dataset using memory bank (a), single-queue MoCo (b) and multi-queue MoCo (c), respectively. }
    \vspace{15pt}
	\label{fig:tsne_visualization}
\end{figure*}

\begin{table}[tbp]
	\centering
	\tiny
    \caption{Performance comparison of the three storage strategies for vectors K on the JHU-Crowd++ dataset. }
    \vspace{5pt}
	\begin{tabular}[]{ccccccc}
		\toprule
		\multirow{2}{*}{Method} & \multicolumn{2}{c}{Normal} & \multicolumn{2}{c}{Adverse} & \multicolumn{2}{c}{Total} \\
		& MAE $\downarrow$ & RMSE $\downarrow$ & MAE $\downarrow$ & RMSE $\downarrow$ & MAE $\downarrow$ & RMSE $\downarrow$ \\
		\midrule
		Memory Bank & 53.7 & 157.6 & 104.9 & 531.1 & 59.8 & 235.7 \\
		Single-queue MoCo & 53.6 & 166.5 & 98.9 & 452.7 & 59.0 & 221.1 \\
		Multi-queue MoCo (Ours) & \textbf{48.1} & \textbf{150.0} & \textbf{82.2} & \textbf{433.5} & \textbf{52.1} & \textbf{205.5} \\
		\bottomrule
	\end{tabular}
	\label{tbl:storage_comparison}
\end{table}

\noindent
\textbf{The storage strategy for vectors K. }
There are three optional storage strategies for vectors K: 
1) memory bank~\cite{wu2018unsupervised};
2) single-queue MoCo~\cite{he2020momentum}; 
3) multi-queue MoCo (ours). 
Extensive experiments are conducted on the JHU-Crowd++ dataset to compare the above strategies. 
The t-SNE~\cite{van2008visualizing} visualization of the vectors Q after the WRL stage on the JHU-Crowd++ dataset is shown in Figure~\ref{fig:tsne_visualization}. 
The memory bank strategy suffers from a severe lack of discriminative capacity.
The representations learned by the single-queue strategy are also not discriminative enough, especially near the rain weather representations in Figure~\ref{fig:tsne_visualization}(b). 
The multi-queue strategy does not suffer from the aforementioned issues. 
The significantly dispersed clusters shown Figure~\ref{fig:tsne_visualization}(c) illustrate that the properties learned from the multi-queue MoCo represent weather information well. 
The performance comparison shown in Table~\ref{tbl:storage_comparison} also demonstrates that the proposed multi-queue MoCo can effectively address the class imbalance problem.
We attribute the phenomenon to the following reasons: 
1) while the memory bank can store a large number of samples with minimal memory consumption, it does not employ a stable strategy to update the encoder, and the sample update frequency is too low, resulting in poor sample consistency;
2) although single-queue MoCo addresses the issue of poor sample consistency by introducing a queue and momentum update strategy, the class imbalance problem in the dataset leads to inconsistencies between the data distribution in the queue and the uniform distribution prior;
3) multi-queue MoCo not only retains the advantages of high sample consistency but also greatly alleviates the problem of inconsistency between data distribution and the uniform prior. 
Thus, it achieves the best performance.

\begin{table}[tbp]
	\centering
	\tiny
    \caption{Ablation study on the JHU-Crowd++ dataset.}
    \vspace{5pt}
	\begin{tabular}[]{cc|cccccc}
		\toprule
		\multirow{2}{*}{WRL} & \multirow{2}{*}{CRR} & \multicolumn{2}{c}{Normal} & \multicolumn{2}{c}{Adverse} & \multicolumn{2}{c}{Total} \\
		& & MAE $\downarrow$ & RMSE $\downarrow$ & MAE $\downarrow$ & RMSE $\downarrow$ & MAE $\downarrow$ & RMSE $\downarrow$ \\
		\midrule
		& & 52.7 & 154.9 & 105.1 & 561.4 & 59.0 & 242.4 \\
		\checkmark & & 49.1 & 152.6 & 87.2 & 458.5 & 53.6 & 213.5 \\
		\checkmark & \checkmark & 48.1 & 150.0 & 82.2 & 433.5 & 52.1 & 205.5 \\ 
		\bottomrule
	\end{tabular}
	\label{tbl:ablation_study}
\end{table}

\noindent
\textbf{Ablation study. }
Ablation studies are performed on the JHU-Crowd++ dataset and the quantitative results are shown in Table~\ref{tbl:ablation_study}. 
We start with the baseline of ConvNeXt-T. 
First, the effectiveness of the WRL stage is tested. 
An improvement of 6.8\%/1.5\% and 17.0\%/18.3\% in MAE/RMSE under normal and adverse weather is achieved compared to the baseline.
From this, it can be seen that most of the performance improvements under adverse weather conditions come from the representation learning in the WRL stage.
This verifies the effectiveness of the proposed multi-queue MoCo for enhancing robustness under adverse weather conditions.
Additionally, the performance improvement under normal weather conditions corroborates the earlier analysis that the strategy that treats different augmentations from the same image as positive samples can aid in strengthening the scene recognition capabilities of the model.
Subsequently, the CRR stage is added, and the best performance is achieved, with a reduction of 2.0\%/1.7\% in MAE/RMSE under normal weather conditions and 5.7\%/5.5\% under adverse weather conditions, respectively.
This demonstrates that the refiner is capable of converting adverse weather representations to the normal domain, enabling the decoder to focus on a single domain.

\begin{table}[tbp]
	\centering
	\tiny
    \caption{Comparison between multi-queue MoCo and other class imbalance contrastive learning strategies on the JHU-Crowd++ dataset.}
    \vspace{5pt}
	\begin{tabular}[]{ccccccc}
		\toprule
		\multirow{2}{*}{Method} & \multicolumn{2}{c}{Normal} & \multicolumn{2}{c}{Adverse} & \multicolumn{2}{c}{Total} \\
		& MAE $\downarrow$ & RMSE $\downarrow$ & MAE $\downarrow$ & RMSE $\downarrow$ & MAE $\downarrow$ & RMSE $\downarrow$ \\
		\midrule
		KCL~\cite{kang2020exploring} (ICLR 20) & 48.7 & 154.3 & 83.9 & 464.6 & 52.9 & 216.1 \\
		BCL~\cite{zhu2022balanced} (CVPR 22) & \textbf{48.1} & 154.3 & 83.2 & 524.2 & 52.3 & 231.9 \\
        TSC~\cite{li2022targeted} (CVPR 22) & 48.8 & 152.1 & \textbf{82.0} & 437.4 & 52.8 & 207.9 \\
        SBCL~\cite{hou2023subclass} (ICCV 23) & 52.0 & 159.2 & 87.1 & 473.0 & 56.2 & 221.4 \\
        \midrule
		Multi-queue MoCo (Ours) & \textbf{48.1} & \textbf{150.0} & 82.2 & \textbf{433.5} & \textbf{52.1} & \textbf{205.5} \\ 
		\bottomrule
	\end{tabular}
	\label{tbl:class_imbalance_comparison}
\end{table}

\begin{table}[tbp]
	\centering
	\tiny
    \caption{Comparison of computational complexity and the number of parameters. The computational complexity is measured by FLOPs when inferencing images with the size of 384 $\times$ 384. 
	}
    \vspace{5pt}
	\begin{tabular}[]{ccc}
		\toprule
		Method & FLOPs & \#param \\
		\midrule
		CLTR~\cite{liang2022end} (ECCV 22) & 37.0G & 43M \\ 
		MAN~\cite{lin2022boosting} (CVPR 22) & 58.2G & 31M \\
		AWCC-Net~\cite{huang2023counting} (ICCV 23) & 58.0G & 30M \\
        Gramformer~\cite{lin2024gramformer} (AAAI 24) & 60.9G & 29M \\
		\midrule
		ConvNeXt-T~\cite{liu2022convnet} (CVPR 22) & 27.0G & 29M \\
		MQCL + ConvNeXt-T & 31.2G & 32M \\
		\bottomrule
	\end{tabular}
	\label{tbl:weight}
\end{table}

\noindent
\textbf{Computation complexity and number of parameters. }
%
%
As shown in Table~\ref{tbl:weight}, compared to the baseline, MQCL introduces only a light-weight refiner, resulting in only 4.2GFLOPs in computational complexity and 3M parameters.
Compared to state-of-the-art methods, MQCL has similar parameter counts to Gramformer and AWCC-Net, while requiring only half their computational cost.
CLTR has higher requirements on both computational and memory cost than MQCL. 
Superior performance to state-of-the-art methods with lower computational costs demonstrates MQCL's successful achievement of light-weight design objective.
\noindent
\textbf{Comparison with other class imbalance strategies. }
In this section, we will discuss the strengths and weaknesses of multi-queue MoCo compared to the five representative imbalanced contrastive learning strategies, and investigate their applicability to our MQCL method.
Attributed to MAK's~\cite{jiang2021improving} strategy of sampling from the real world, the diversity of tail-end classes is significantly increased, the issue of class imbalance can be effectively addressed.
However, it is not applicable to our crowd counting task because only unlabeled samples can be supplemented, but in crowd counting, head position labels are required for learning. 
The rest four methods can be applied to our method, we conduct experiments on the JHU-Crowd++ dataset to compare and analyze their performance results.
In the experiment setup, only $\mathcal{L}_{contra1}$ in the MQCL is replaced with the loss functions of the four imbalanced contrastive learning strategies, and all other settings remains unchanged. 
The results are shown in Table~\ref{tbl:class_imbalance_comparison}. 
The performance of KCL~\cite{kang2020exploring} is worse than multi-queue MoCo because KCL samples a sufficient number of positive samples, but the negative samples remain imbalanced.
SBCL~\cite{hou2023subclass} fundamentally alleviates the issue of class imbalance through the method of subclass partitioning.
The experiments in~\cite{hou2023subclass} show that SBCL outperforms KCL~\cite{kang2020exploring} and TSC~\cite{li2022targeted} in image classification tasks.
However, it achieved the lowest performance in this experiment.
We think the approach of dividing the normal weather class into multiple subclasses may bring confusion to the CRR stage.
Because in the CRR stage, the representations are pulled towards the normal weather domain, which, however, has multiple clusters.
BCL~\cite{zhu2022balanced} combines gradient balancing and batch data balancing;
TSC~\cite{li2022targeted} introduces cluster center priors.
They achieved performance almost as excellent as multi-queue MoCo.
Although these three methods exhibit similar performance, they address the imbalance problem from different perspectives, each with its own significance.

\section{Conclusion and Limitations}
In this paper, we propose a contrastive learning-based method called MQCL to tackle the problem of class-imbalanced adverse weather crowd counting. 
The core contributions of this paper are multi-queue MoCo and the contrastive-learning-based refining method, which respectively achieve weather-aware representation learning and representation conversion, addressing the dual challenges of image degradation and class imbalance.
Through the proposal of MQCL, we aim to provide a novel multi-domain perspective for addressing the crowd counting task.
Nevertheless, our work still has limitations.
For example, our focus is on the application of multi-queue MoCo in the crowd counting task.
But it is a generic method to deal with data imbalance and holds significant reference value for other domains. 
Exploring its potential in more tasks is a direction for our future work.
%


\bibliography{mybibfile}

\end{document}